\documentclass[10pt,twocolumn,letterpaper]{article}

\usepackage{iccv}
\usepackage{times}
\usepackage{epsfig}
\usepackage{graphicx}
\usepackage{amsmath}
\usepackage{amssymb}

\usepackage{booktabs}
\usepackage{url}

\usepackage[pagebackref=true,breaklinks=true,letterpaper=true,colorlinks,bookmarks=false]{hyperref}

\iccvfinalcopy 

\def\iccvPaperID{xxxx}

\ificcvfinal\fi

\makeatletter
\@namedef{ver@everyshi.sty}{}
\makeatother
\usepackage{graphicx}
\usepackage{pgfplots}
\usepackage{pgfplotstable}
\pgfplotsset{compat=newest}
\usepackage{siunitx}

\usepackage[capitalize]{cleveref}
\crefname{section}{Sec.}{Secs.}
\Crefname{section}{Section}{Sections}
\Crefname{table}{Table}{Tables}
\crefname{table}{Tab.}{Tabs.}

\usepackage{multirow}
\usepackage{xcolor,colortbl}

\usepackage{enumitem,bigdelim}
\usepackage{lipsum}
\usepackage{fontawesome}  

\newcommand{\boldparagraph}[1]{\vspace{0.1em}\noindent{\bf #1} }

\newcommand{\obj}{\mathcal{P}}
\newcommand{\tex}{\mathcal{T}}
\newcommand{\Ren}{\mathcal{R}}
\newcommand{\mot}{\mathcal{N}}

\newcommand{\I}{I}

\newcommand{\Loss}{\mathcal{L}}
\newcommand{\Feat}{F}

\newcommand{\param}{\alpha}

\newcommand{\lock}{\text{\faLock}}

\newcommand\blfootnote[1]{%
  \begingroup
  \renewcommand\thefootnote{}\footnote{#1}%
  \addtocounter{footnote}{-1}%
  \endgroup
}
\newcommand{\addimg}[1]{\includegraphics[width=0.12\linewidth]{#1}}

\begin{document}

\title{Tracking by 3D Model Estimation of Unknown Objects in Videos}

\author{Denys Rozumnyi$^{1,4}$
\hspace{0.5cm}
Ji\v{r}\'{i} Matas$^{4}$
\hspace{0.5cm}
Marc Pollefeys$^{1}$
\hspace{0.5cm}
Vittorio Ferrari$^{3}$
\hspace{0.5cm}
Martin R. Oswald$^{1,2}$\\[0.0em]
$^{1}$Department of Computer Science, ETH Zurich
\hspace{0.5cm} 
$^{2}$University of Amsterdam\\
$^{3}$Google Research
\hspace{0.5cm}
$^{4}$Czech Technical University in Prague
\\[0.0em]
}
\maketitle
\ificcvfinal\thispagestyle{empty}\fi

\begin{abstract}
Most model-free visual object tracking methods formulate the tracking task as object location estimation given by a 2D segmentation or a bounding box in each video frame.
We argue that this representation is limited and instead propose to guide and improve 2D tracking with an explicit object representation, namely the textured 3D shape and 6DoF pose in each video frame.
Our representation tackles a complex long-term dense correspondence problem between all 3D points on the object for all video frames, including frames where some points are invisible.
To achieve that, the estimation is driven by re-rendering the input video frames as well as possible through differentiable rendering, which has not been used for tracking before.
The proposed optimization minimizes a novel loss function to estimate the best 3D shape, texture, and 6DoF pose.
We improve the state-of-the-art in 2D segmentation tracking on three different datasets with mostly rigid objects.
\end{abstract}

\section{Introduction} \label{sec:intro}

Visual object tracking is an important task in computer vision.
A wide range of challenges~\cite{vot2020,vot2021}, benchmarks~\cite{VOT_TPAMI}, and methods~\cite{CSR,D3S,ostrack} have been proposed in recent years.
Given an input video and an annotation of the object in the first frame, tracking methods localize the object with a 2D segmentation (or a 2D bounding box) in all following frames.
However, they do not attempt to reconstruct the 3D shape and its 6DoF motion and settle for 2D segmentation as a low-level representation of the object state.
On the one hand, this makes those methods general in terms of prior assumptions on object shape and motion.
On the other hand, the information extracted by the segmentation is limited, \eg to whole object editing, and does not support many applications.

In this paper, we propose a generic approach for tracking that is fundamentally different from what standard trackers perform~\cite{VOT_TPAMI,vot2021}.
Instead of a mere 2D segmentation, we estimate the latent object 3D shape, texture, and 6DoF pose in each video frame to improve 2D tracking.
This formulation of tracking solves a more complex long-term dense correspondence problem between all 3D points on the object, even of points that are not visible (Fig.~\ref{fig:teaser}, colored lines).
This is useful in many applications,~\eg augmented reality, motion interpretation and analysis, point matching, per-pixel editing, robotics, object manipulation and grasping. 

\newcommand{\AlgName}[1]{\rotatebox{90}{\scriptsize #1}}
\setlength{\fboxsep}{0pt}

\newcommand{\addimgTno}[1]{\includegraphics[width=0.134\linewidth]{#1}}
\newcommand{\addimgT}[1]{\framebox{\addimgTno{#1}}}
\newcommand{\makeOneRowTno}[7]{\addimgTno{#1#2} &\addimgTno{#1#3} & \addimgTno{#1#4} &\addimgTno{#1#5} & \addimgTno{#1#6}  & \addimgTno{#1#7}  }
\newcommand{\makeOneRowT}[7]{\addimgT{#1#2} &\addimgT{#1#3} & \addimgT{#1#4} &\addimgT{#1#5} & \addimgT{#1#6}  & \addimgT{#1#7}  }

\newcommand{\makeRowT}[7]{%
\AlgName{\phantom{Iy} Input} & \makeOneRowTno{#1is}{#2}{#3}{#4}{#5}{#6}{#7} \\
\AlgName{\phantom{Iy} Ours} & \makeOneRowT{#1f}{#2}{#3}{#4}{#5}{#6}{#7} & \addimgTno{#1tex_deep} \\
\AlgName{RGB proj.} & \makeOneRowT{#1l_r}{#2}{#3}{#4}{#5}{#6}{#7} & \addimgTno{#1l_tex} \\
}

\newcommand{\mc}[2]{\multicolumn{#1}{c}{#2}}
\definecolor{LightCyan}{rgb}{0,0.99,1}
\definecolor{DarkGreen}{rgb}{0.47,0.72,0.28}
\definecolor{DarkRed}{rgb}{0.98,0.2,0.11}
\newcommand{\linesize}{\tiny}

\begin{figure}
\centering
\small
\setlength{\tabcolsep}{0.0em} 
\renewcommand{\arraystretch}{0} 
\begin{tabular}{@{}c@{\hskip 0.1em}cccccc@{\hskip 0.1em}c@{}}
\makeRowT{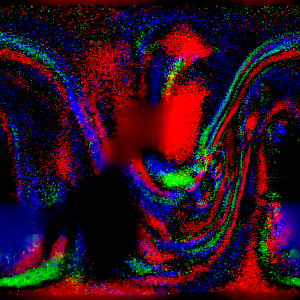}{2}{4}{5}{6}{8}{0}\addlinespace[0.1em]
 \multirow{3}{*}{\AlgName{vis}} & \mc{2}{\cellcolor{DarkGreen!100} \phantom{\linesize x}} & & & & \mc{1}{\cellcolor{DarkGreen!100} \phantom{\linesize x}} \\
 & & \mc{4}{\cellcolor{LightCyan!100} \phantom{\linesize x}} & \\
 & & & & \mc{2}{\cellcolor{DarkRed!100} \phantom{\linesize x}} & \\\addlinespace[0.1em]
 & \multicolumn{6}{c}{Sequence} & Texture \\
\end{tabular}
\vspace{-0.5em}
\caption{\textbf{Improving 2D tracking with explicit 3D modeling.} Starting from initial segmentations given by a tracker~\cite{ostrack} (top row), we jointly optimize all object parameters to re-render the input video as close as possible. 
Long-term correspondences are visualized by colored points and their visibility by colored lines underneath.  
In contrast, the standard tracker reports essentially no motion (just slightly changing 2D segmentation).
We use S2DNet features~\cite{S2DNet} for reprojection error (middle row) and back-project RGB values only for visualization (bottom row). }
\label{fig:teaser}
\end{figure}

The core of the approach is to find the object shape, pose, and texture -- the parameters that define the object -- whose reprojection is most consistent with input video frames.
The parameter estimation is driven by re-rendering the input video via differentiable rendering instead of directly predicting the 2D segmentation in the next frame based on features extracted in the previous frames.
This modern differentiable rendering way of thinking was demonstrated to be successful in various applications, such as 3D reconstruction~\cite{softras,dibr}, structure-from-motion~\cite{pixsfm}, reconstruction of clothed humans~\cite{Huang_2020_CVPR}, and deblurring of motion-blurred objects~\cite{sfb,mfb}, but never in tracking.
To initialize our optimization process, we use segmentations given by a standard 2D tracker.
We demonstrate experimentally that the proposed tracking method improves segmentation accuracy of this initial tracker.

Another related line of research is multi-view 3D object reconstruction~\cite{3dr2n2,nerf,pixelnerf}.
Those methods solve the task with given additional information, usually with camera poses (or alternatively object poses), camera calibration matrix, depth information, object segmentation, \etc.
Moreover, these methods do not process the sequences causally and reconstruct the whole scene at once.
Most methods assume either a perfect segmentation or objects in front of a blank background~\cite{3dr2n2,sharf}.
They usually require distinctive texture and are trained on large datasets, in some cases even for a particular object class~\cite{sharf}.

A third group, methods for 6DoF pose estimation~\cite{hodan2018bop, hodan2020bop, labbe2020}, require another type of additional information as input -- the 3D shape of the object appearing in the video~\cite{hodan2018bop} (or a small set of 3D shapes~\cite{lipson2022coupled, labbe2020}).
Some methods reconstruct higher than 6DoF poses~\cite{you2022cppf} or poses of articulated objects~\cite{smpl}.
In contrast, our method requires no additional information and only uses RGB information as input.
To summarize, we make the following contributions:
\begin{enumerate}[itemsep=0.1pt,topsep=3pt,leftmargin=*,label=\textbf{(\arabic*)}]
    \item we propose the first model-free object tracking method that explicitly estimates the latent object 3D shape, texture, and 6DoF motion from an RGB sequence.  The tracker is aware of which part of its surface is visible,
    \item we introduce differentiable rendering to tracking to optimize object parameters to re-render the input video,
    \item we introduce deep surface textures to improve tracking robustness under view-dependent illumination changes,
    \item experiments demonstrate improved tracking accuracy compared to the initial tracker, more accurate 2D segmentation, and dense long-term correspondences on the object surface, even after self-occlusion. It sets new state-of-the-art results on tracking datasets with rigid objects such as CDTB (part of VOT challenge), coin tracking, and In-hand Object Manipulation.
\end{enumerate}

\section{Related work}  \label{sec:related}
\boldparagraph{Visual object tracking in 2D.}
Visual object tracking is one of the oldest tasks in computer vision dating back to the Lucas-Kanade tracker~\cite{klt,klt2}.
Discriminative Correlation Filters have been successfully used for the tracking task~\cite{kcf,CSR}.
More recently, methods based on deep learning have been developed~\cite{atom,D3S,ostrack}.
Most trackers localize the object only in 2D by a 2D bounding box or a 2D segmentation.
In contrast, we aim to improve tracking by explicitly recovering rough 3D shape, texture, and 6DoF pose.

\boldparagraph{3D reconstruction.}
In recent years, 3D reconstruction has gained significant attention.
Structure-from-Motion pipelines, such as COLMAP~\cite{schoenberger2016sfm}, have become widely used.
Other methods are based on Neural Radiance Fields~\cite{nerf,pixelnerf}, Transformers~\cite{Wang_2021_ICCV}, Recurrent Neural Networks~\cite{3dr2n2}.
Methods designed specifically for 3D object reconstruction have also been proposed~\cite{Wang_2021_ICCV,sharf,3dr2n2}
However, they always assume a perfect segmentation or trivial backgrounds and impose strong priors on object shape~\cite{sharf}.
Similarly, BackFlow~\cite{Wang2019} reconstructs simple objects that are manipulated by human hands in RGB-D videos (we use only RGB).
Moreover, they assume that a user provides scribbles to segment out the background in the first frame. 

\boldparagraph{3D and 6DoF object pose estimation or tracking.}
An approach to enhance 2D tracking with 3D reconstruction was proposed by~\cite{Kart_2019_CVPR}, but it requires depth information as input and estimates latent 3D representation as a point cloud by concatenating RGB-D segmentations, without any optimization, viewpoint consistency, and explicit surface modeling. 
Thus, it looses many desirable properties, \eg self-occlusions awareness and long-term point-to-point correspondence. 
The field of 6DoF object pose estimation is vast. 
A number of challenges and benchmarks have been proposed~\cite{hodan2018bop,hodan2020bop}, comprising of various methods~\cite{labbe2020,epos,wang2020self6d}.
However, the main limiting factor of these methods is that they need the known 3D shape of the object as input.
Some methods were proposed for 6D tracking of unknown objects~\cite{7139520,Leeb2019MotionNets6T,gladkova2022directtracker}.
In contrast to our approach, these methods do not estimate the 3D shape but only try to predict 6DoF object pose given initial object segmentation. 

\boldparagraph{Differentiable rendering.}
The classical computer graphics rendering pipeline was recently made differentiable in various ways.
Most methods are based on soft rasterization~\cite{softras,dibr,dibrpp}, differentiable sphere tracing~\cite{liu2020dist}, neural radiance fields (NeRF)~\cite{nerf}, \etc.
NeRF was also combined with pose estimation,~\eg~\cite{inerf,barf}.
Differentiable rendering techniques were also used in various applications: 3D reconstruction~\cite{softras,dibr}, monocular 6DoF pose estimation~\cite{wang2020self6d}, reconstruction of clothed humans~\cite{Huang_2020_CVPR}, and deblurring of motion-blurred objects~\cite{sfb,mfb}.
These methods are related to our work in their use of differentiable rendering, but we apply it to a different task.

As a consequence, the proposed method is the first to use differentiable rendering in the object tracking task in videos with joint 3D reconstruction and 6DoF pose estimation of unknown objects in the wild.

\section{Method}  \label{sec:method}
%
\begin{figure*}[t]
\centering
\includegraphics[width=\linewidth]{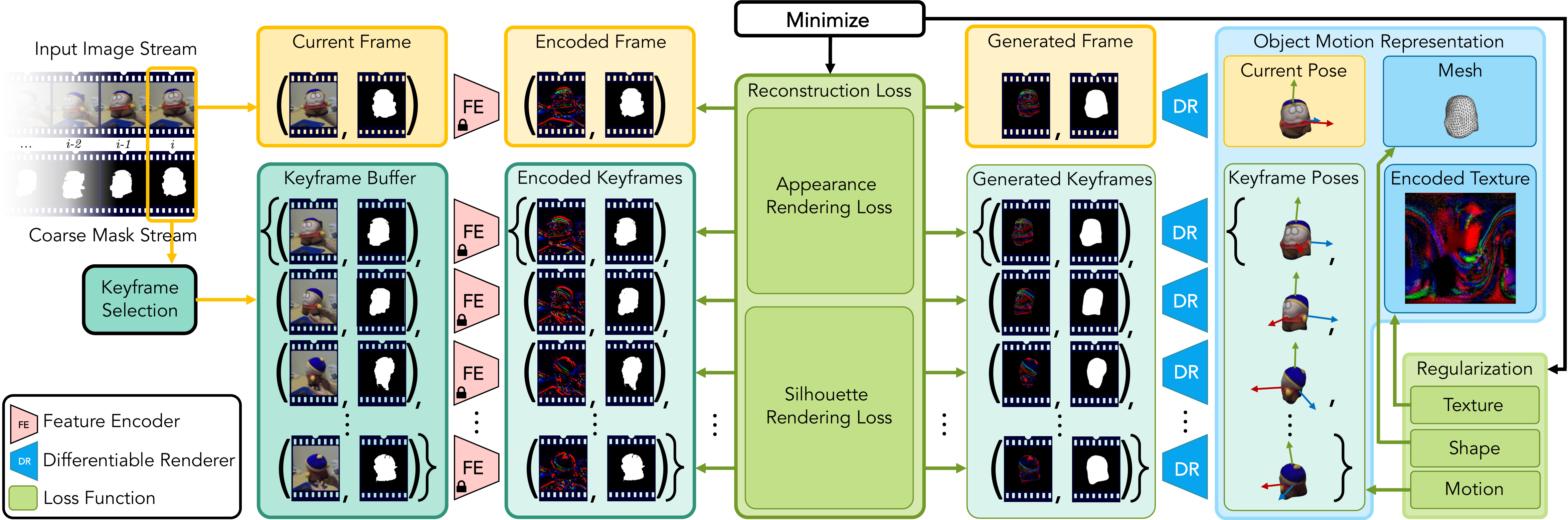}
\caption{\textbf{Method Overview. Left-to-middle:} The proposed method processes a stream of images and coarse masks and jointly estimates 3D object shape, texture, and 6DoF poses. As a first step, the input images are transferred using a feature encoder (\textbf{FE}) to better handle view-dependent brightness changes. The feature encoder is pre-trained and remains unchanged in our optimization (as indicated by the \lock-symbol). The feature encoder is optional and can be replaced with an identity map in which case our method computes the reconstruction loss in RGB space rather than the more general feature space,~\eg S2DNet~\cite{S2DNet}. Concurrently, we collect keyframes that serve as reference information for object motion estimation.
\textbf{Right-to-middle:} The  method can be seen as a generative approach: Starting from an object shape (mesh), texture, and 6DoF object poses, we generate images and corresponding masks with a differentiable renderer (\textbf{DR}). At test time, the inverse problem is solved by refining the object motion representation to minimize the reconstruction loss. Further regularization on the object shape, texture, and motion further stabilizes the optimization and helps avoiding undesired local minima.
}
\label{fig:pipeline_overview}
\end{figure*}

We assume that the input is a video stream $\{\I_1,\dots,\I_N\}$ of $N$ frames and the position of the object in the first frame, given either by a segmentation or a bounding box.
As the first step of our method, we approximately segment the object with a state-of-the-art 2D tracker,~\eg~\cite{ostrack,alpharefine}, resulting in a binary segmentation per frame $\{M_1,\dots,M_N\}$.
The desired output of our method is improved binary segmentation and additional latent object representation, which is a textured 3D shape of the tracked object and its 6DoF pose in each frame.
We obtain improved segmentations by rendering the estimated 3D object at the estimated 6DoF pose in each frame.
The improvement comes from seeking for a 3D shape, texture, and 6DoF poses that re-render the input video frames as accurately as possible. We optimize over the space of 3D shapes, textures, and 6DoF poses with differentiable rendering.

The whole pipeline (Fig.~\ref{fig:pipeline_overview}) is designed in such a way that the initial segmentations only guide the object's 6DoF tracking and 3D reconstruction without being overly dependent on their accuracy.
We run the optimization sequentially for each frame in an online manner making the method causal,~\ie depending only on the past.

In principle, at each new frame we could optimize the object 3D shape, texture, and 6DoF poses over all past video frames.
However, for efficiency we maintain a subset of so-called \textit{keyframes} indexed by $K_n \subseteq [1,\dots,n-1]$, which act as sparse anchors on which the optimization is performed.
The main idea is that each keyframe provides a significantly different view of the object.
The subset of keyframes is sequentially updated every time the process moves to a new frame, based on criteria described in Sec.~\ref{sec:key}.

\subsection{Representations}
\boldparagraph{Object shape representation.}
We estimate a single 3D object shape that is fixed throughout the video.
The object is represented in its canonical space by a textured triangular mesh with a set of vertices, faces, and texture mapping.
The shape parameters $\obj$ are vertex offsets from a prototype mesh, which deform it into the actual shape of the object in the video.
The mesh topology and faces are fixed. 

\boldparagraph{Object appearance representation.}
The appearance of the object is represented by a texture map of features $\tex$.
The function that maps RGB values to a feature representation is denoted by $\Feat(\cdot)$.
In the simplest case, $\Feat$ can be an identity mapping.
Then, the features are simply RGB values. 
However, in most cases, when the object moves, the RGB values change due to lighting effects, shadows, illumination changes, \etc. 
To overcome this problem, invariant features have been studied in image matching and structure-from-motion,~\eg~\cite{pixsfm}.
We use a similar approach and extract 128-dimensional S2DNet~\cite{S2DNet} features, leading to a deep surface texture.
Texture mapping is a function that denotes the correspondences between pixels in the texture map and mesh faces, and it is fixed throughout the optimization.

\boldparagraph{Object motion representation.}
The object motion $\mot$ is a temporal sequence of 6DoF poses.
The 6DoF pose $\mot_n = (T_n, Q_n)$ in each frame $n$ is represented by 3D translation and 3D rotation.
Translations are represented by 3D vectors $T_n$ and rotations by normalized 4D quaternions $Q_n$.

\newcommand{\addimgv}[1]{\includegraphics[width=0.109\linewidth]{#1}}

\newcommand{\makeOneRowv}[1]{\addimgv{#10} & \addimgv{#11} & \addimgv{#12}  &\addimgv{#13}  &\addimgv{#14}  & \addimgv{#15}  & \addimgv{#16}  & \addimgv{#18} }
\newcommand{\makeRowv}[1]{%
\AlgName{\phantom{Inputt} Input} & \makeOneRowv{#1i} \\
\AlgName{\phantom{Inputut} Ours} & \makeOneRowv{#1f} & \addimgv{#1tex_deep}\\
\AlgName{\phantom{Iyyy} RGB proj.} & \makeOneRowv{#1l_r} & \addimgv{#1l_tex}\\
}

\newcommand{\makeline}[2]{\mc{#1}{\cellcolor{#2!100} \phantom{\linesize x}}}

\begin{figure*}
\centering
\small
\setlength{\tabcolsep}{0.0em} 
\renewcommand{\arraystretch}{0} 
\begin{tabular}{@{}c@{\hskip 0.1em}cccccccc@{\hskip 0.1em}c@{\hskip 1em}c@{}}

\makeRowv{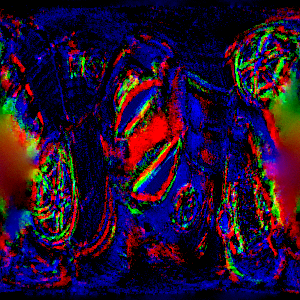}
\multirow{4}{*}{\AlgName{vis}} & \mc{2}{} & \makeline{3}{orange} \\
 & \makeline{1}{DarkGreen} & \mc{5}{} & \makeline{2}{DarkGreen} \\
 \mc{5}{} & \makeline{3}{LightCyan}    \\
 & \makeline{3}{yellow}   \\
 & \multicolumn{8}{c}{Keyframes} & Texture \\
\end{tabular}
\vspace{-0.5em}
\caption{\textbf{Evaluation on a vacuum cleaner sequence.} This example highlights the need for more robust features like S2DNet~\cite{S2DNet} (middle row) that are robust to various illumination changes. RGB color values vary significantly within the input sequence (top row), and directly optimizing RGB texture leads to failures. In contrast, we back-project RGB features (bottom row) after optimization has finished. Long-term correspondences are marked with colored points and their visibility with colored lines. 
Texture visualization is a UV map~\cite{blender}.
}
\label{fig:vacuum}
\end{figure*}

\subsection{Loss terms}
%
\boldparagraph{Appearance and silhouette rendering.} \label{sec:loss}
Let us define a rendering function $\Ren$ that takes three inputs, object shape $\obj$, object appearance $\tex$, and 6DoF pose ($T_n, Q_n$), and renders the object at the given pose.
This function provides two outputs, namely object appearance $\Ren_F(\obj, \tex, \mot_n)$ and object silhouette $\Ren_S(\obj, \tex, \mot_n)$ (subscript $F$ for 'Features', and $S$ for 'Silhouette').
We assume that the camera is fixed.
The camera projection matrix is fixed to $45^{\circ}$ field-of-view angle.
Given the rendered appearance and silhouette, we define below loss terms for every frame $n$.

\boldparagraph{Appearance rendering loss.}
This loss measures how well the estimated 3D shape and texture re-renders the input video frames.
This re-rendering is the main driving force of the proposed approach.
Concretely, we compare the object appearance features rendered at the given pose to the observed video features $\Feat(\I_i)$. 
We use the Cauchy loss $\gamma$ with a scale of 0.25 as in~\cite{pixsfm}:
\begin{small}
\begin{equation}
\label{eq:feat}
\Loss_{\Feat} = \mu_n \!\!\!\!\!\! \sum_{i \in \{K_n \cup n\}}  \left\| \Ren_S(\obj, \tex, \mot_i) \cdot \Big( \Ren_F(\obj, \tex, \mot_i) - \Feat(\I_i) \Big) \right\|_{\gamma} 
\end{equation}
\end{small}
Here, the summation runs over the current set of frames $K_n$ and the current frame $n$.
The factor $\mu_n = (|K_n|+1)^{-1}$ normalizes the loss.

\boldparagraph{Silhouette rendering loss.}
This loss is based on the initial segmentations of the 2D tracker we start from.
This loss keeps the method anchored to the correct object since there might be multiple objects in the scene.
We define it as a sum of two terms: 
\begin{equation}
\label{eq:sil}
\begin{split}
\Loss_S = \mu_n \sum_{i \in \{K_n \cup n\}} \bigg( \Big( 1 - \text{IoU}\big(M_i, \Ren_S(\obj, \tex, \mot_i)\big) \Big) + \\
 + \|\text{DT}(M_i) \cdot \Ren_S(\obj, \tex, \mot_i)\| \bigg) \, . 
\end{split}
\end{equation}
The first term is the intersection-over-union (IoU) between the rendered silhouette of the estimated 3D shape and the initial segmentation $M_i$.
When the rendered silhouette and the initial segmentation are near, this loss behaves well.
However, when they are far away, this term is zero, leading to zero gradient flow during optimization (Sec.~\ref{sec:optim}).
To address this case, the second term in~\eqref{eq:sil} attracts the object silhouette to move closer to the initial segmentation.
It measures the average distance of each pixel of the silhouette to the closest point in the initial segmentation $M_i$.
It is efficiently implemented by computing a distance transform (DT) image of the initial segmentation and multiplying it by the rendered silhouette.

\boldparagraph{Motion regularization.}
To encourage motion smoothness, we penalize the difference between the translation $T_n$ in the current frame and the latest previous keyframe.
Similarly, we penalize the angular difference in rotations $Q_n$:
\begin{equation}
\label{eq:motion}
\begin{split}
\Loss_M = \max\Big(0, \frac{1}{K_{n}[-1]-n} \| T_{K_{n}[-1]} - T_{n} \|_2 - \nu_T \Big) + \\
 + \max\Big(0, \frac{1}{K_{n}[-1]-n} \measuredangle (Q_{K_{n}[-1]}, Q_{n}) - \nu_Q \Big)  \, ,
\end{split}
\end{equation}
where $K_{n}[-1]$ denotes the index of the most recent keyframe before the current frame.
We do not penalize differences of rotations and translations to the last keyframe smaller than thresholds $\nu_T = 0.1$ and $\nu_T = 30^{\circ}$, as we consider these motions to be in a normal range.
Effectively, the loss penalizes drastic pose changes w.r.t.~the last keyframe.

\boldparagraph{Shape regularization.}
To promote 3D shape smoothness, we use the Laplacian loss $\Loss_L$ as defined in~\cite{softras}.

\boldparagraph{Texture regularization.}
For a smooth texture map, we use a total variation loss $\Loss_T$ that computes the average appearance difference between neighboring pixels in the texture map (using $\Loss_{\Feat}$ to measure the difference).

\boldparagraph{Total loss.}
The total loss is a weighted sum of all the above loss terms: $\Loss = \param_{\Feat} \Loss_{\Feat} + \param_S \Loss_S + \param_M \Loss_M + \param_L \Loss_L + \param_T \Loss_T$. In our experiments, we set $\param_{\Feat} = 1$, $\param_S = 1$, $\param_M = 1$, $\param_L = 1000$, and $\param_T = 0.001$.

\subsection{Optimization}\label{sec:optim}
%
To minimize the total loss $\Loss$ in each frame, we use the ADAM~\cite{adam} optimizer with a learning rate of $0.1$.
The optimization is stopped early when the appearance rendering loss reaches $\Loss_{\Feat} < \tau_{\Feat}$, where $\tau_{\Feat}$ is the convergence threshold.
For RGB features we use $\tau_{\Feat} = 0.2$, while for S2DNet~\cite{S2DNet} features we use $\tau_{\Feat} = 0.05$.
If this stopping criterion is not reached within $500$ iterations, we declare the current frame a failure.
This automatic self-assessment declares that the initial segmentation cannot be improved by the proposed method.
Failure can happen due to various reasons,~\eg bad initial segmentation, difficult object motion, non-rigid object deformation (we assume rigid object), extreme lighting or texture changes on the object itself in the real world (we assume constant texture throughout the sequence).
The self-assessment is especially helpful when the method is evaluated on large datasets, where a rigid object in the scene cannot be guaranteed.
When a failure is detected, our method automatically degenerates to returning the initial segmentation for that frame. This allows our method to operate in more general conditions, improving over the initial tracker when it can and simply returning its output when it cannot.
In our experiments, self-assessment detects a failure in $1$-$5\%$ of frames depending on the dataset.

\boldparagraph{Keyframe selection criteria.}\label{sec:key}
Since the optimization is anchored to the subset of keyframes, they have to be carefully selected.
The subset of keyframes is dynamic, as it changes over time.
Selection criteria are re-evaluated after the optimization has finished for each frame. 
Thus, keyframes can fall out of the subset when one of the selection criteria do not hold anymore.
We have four selection criteria for the current frame to become a keyframe:
\begin{enumerate}[itemsep=0.1pt,topsep=3pt,leftmargin=*,label={(\arabic*)}]
\item The rendered object silhouette aligns well with the 2D segmentations,~\ie $\Loss_S < 0.3$.
\item The rendered object appearance is consistent with the input frame,~\ie $\Loss_{\Feat} < \tau_{\Feat}$.
\item The current frame and the latest keyframe provide significantly different views of the object,~\ie relative translation is greater than half the object size or relative rotation is greater than $45^{\circ}$.
\item Finally, the number of keyframes is restricted to a maximum $\param_K$ frames; thus, only the most recent $\param_K$ are kept (discussed in Sec.~\ref{sec:ablation}).
\end{enumerate}

\boldparagraph{Differentiable rendering.}
As part of the forward model, we render object appearance and silhouette.
To make this step differentiable, we apply differentiable rendering proposed in~\cite{dibr}.

\boldparagraph{Initialization.}
The object 3D shape is initialized to a prototype mesh, which is a sphere with 1212 vertices.
Vertex offsets are initialized to zero. 
The texture map has dimension $300 \times 300$ and is initialized to zero values.
The 6DoF pose in the first frame is initialized to 2D location in the center of the image, depth such that the object covers the whole image, and zero 3D rotation.
Every next 6DoF pose is initialized to the pose estimated in the previous frame.

\boldparagraph{Runtime.}
Currently, the method runs at $2$ sec.~per frame on average.
However, the runtime can be improved by simpler shape representations, better motion prediction, and hierarchical optimization.  
Moreover, the method is highly parallelizable,~\eg all keyframes can be rendered in parallel.

\setlength{\fboxsep}{0pt}
\newcommand{\addimgCNoframe}[1]{{\includegraphics[width=0.145\linewidth]{#1}}}

\newcommand{\addimgC}[1]{\framebox{\includegraphics[width=0.145\linewidth]{#1}}}

\newcommand{\makeOneRowCNoframe}[7]{\addimgCNoframe{#1#2} &\addimgCNoframe{#1#3} & \addimgCNoframe{#1#4} &\addimgCNoframe{#1#5} & \addimgCNoframe{#1#6}  & \addimgCNoframe{#1#7}   }

\newcommand{\makeOneRowC}[7]{\addimgC{#1#2} &\addimgC{#1#3} & \addimgC{#1#4} &\addimgC{#1#5} & \addimgC{#1#6}  & \addimgC{#1#7}   }

\newcommand{\makeRowC}[7]{%
\AlgName{\phantom{Inputt} Input} & \makeOneRowCNoframe{#1i}{#2}{#3}{#4}{#5}{#6}{#7} \\
\AlgName{\phantom{Input} Segm.~\cite{ostrack}} & \makeOneRowC{#1s}{#2}{#3}{#4}{#5}{#6}{#7} \\
\AlgName{\phantom{OursOur} Ours} & \makeOneRowC{#1r}{#2}{#3}{#4}{#5}{#6}{#7} \\
}

\begin{figure*}
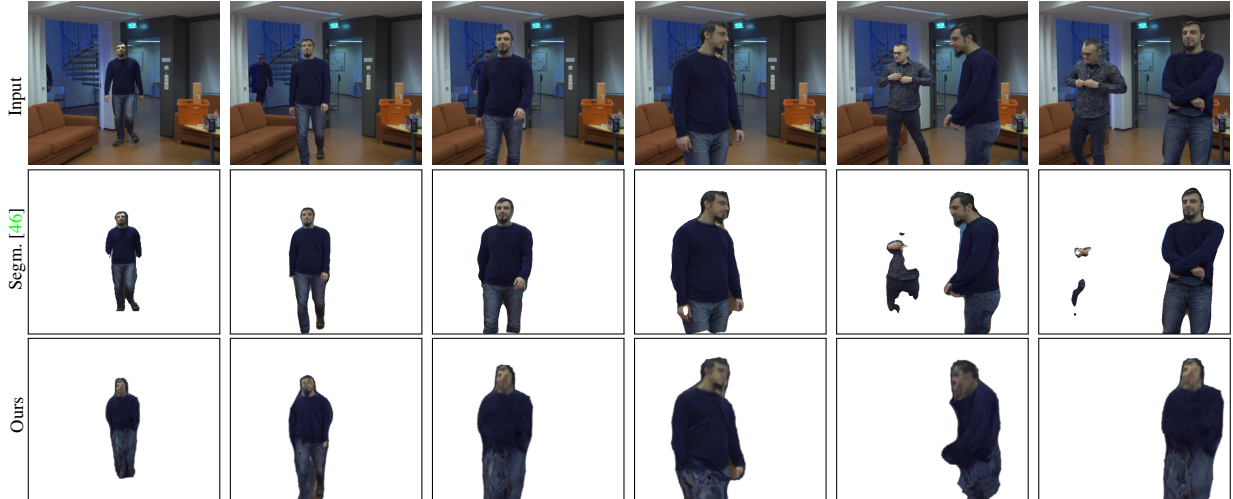

\centering
\small
\setlength{\tabcolsep}{0.2em} 
\renewcommand{\arraystretch}{0.5} 
\begin{tabular}{@{}c@{\hskip 0.1em}cccccc@{}}

\makeRowC{imgs/cdtb_human/}{0}{1}{2}{3}{5}{6}

\end{tabular}
\caption{\textbf{Results with slight non-rigid deformations.} Our method successfully reconstructed and improved the segmentation accuracy (by $11.1 \%$ IoU on average) on a human body sequence from the CDTB dataset~\cite{cdtb}. }
\label{fig:cdtb}
\end{figure*}

\newcommand{\winner}[1]{\textbf{#1}}
\begin{table}
\centering
\small
\setlength{\tabcolsep}{15pt} 
\newcommand{\MySkip}{\hskip 1.3em}  
\begin{tabular}{lcc}
\toprule
Method & IoU & $\Delta$  \\
\midrule
CSR-DCF~\cite{CSR} & 0.498  & - \\
CSR-DCF~\cite{CSR} + Ours  & 0.512 & 2.8 \% \\
\midrule
D3S~\cite{D3S} & 0.732  & - \\
D3S~\cite{D3S} + Ours  &  0.765 & 4.5 \% \\
\midrule
OSTrack~\cite{ostrack} & 0.758 & - \\
OSTrack~\cite{ostrack} + Ours  & 0.788 & 4.0  \% \\
\bottomrule
\end{tabular}
\vspace{0.3em}
\caption{\textbf{Segmentation accuracy on the CDTB dataset~\cite{cdtb}.} Our method improves initial segmentations from trackers in all cases. }
\label{tab:cdtb}
\end{table}

\section{Evaluation}  \label{sec:eval}

The first step of the proposed method is running a state-of-the-art 2D tracker to obtain initial segmentation masks (by default OSTrack~\cite{ostrack} with AlphaRefine~\cite{alpharefine}, the winner of VOT-ST 2022 challenge~\cite{Kristan2022a}).
Hence, we evaluate the improvement in 2D segmentation brought by our method by rendering the silhouette of the reconstructed object on the video frames.

Our method assumes the object is rigid.
Most tracking datasets, \eg~\cite{VOT_TPAMI,got10k}, contain non-rigid objects such as humans and animals.
Therefore, we selected datasets with mostly rigid objects. 
First, we evaluate on the large CDTB~\cite{cdtb} dataset that contains mostly rigid objects (part of the VOT dataset collection~\cite{VOT_TPAMI}).
Second, we evaluate on two smaller datasets, the Coin Tracking dataset~\cite{coin} and the in-hand object manipulation dataset~\cite{Wang2019}, both of which contain only rigid objects.

\begin{table}
\centering
\small
\setlength{\tabcolsep}{15pt} 
\newcommand{\MySkip}{\hskip 1.3em}  
\begin{tabular}{llc}
\toprule
 Method & Input & IoU \\
\midrule
OSVOS~\cite{osvos} & RGB & 87.97 \% \\
OSTrack~\cite{ostrack} & RGB & 93.26 \% \\
Ours (OSTrack init) & RGB &  \textbf{94.05 \%} \\
BackFlow~\cite{Wang2019} & RGB-D & 93.52 \% \\
\bottomrule
\end{tabular}
\vspace{0.3em}
\caption{\textbf{In-hand object manipulation dataset~\cite{Wang2019}.} Our method achieves the highest segmentation accuracy. }
\label{tab:inhand}
\end{table}

\subsection{CDTB dataset}
%
The "Color-and-Depth general visual object
Tracking Benchmark" (CDTB)~\cite{cdtb} is part of the VOT-RGBD challenge suite~\cite{VOT_TPAMI}. It is the only standard tracking dataset that contains mostly rigid objects.
Since we do not require depth information, we drop the depth channel.
The dataset contains 80 sequences with an average of 1274 frames each.

To test how much the initial 2D tracker influences the final results of our method, we use three modern 2D trackers: D3S~\cite{D3S}, CSR-DCF~\cite{CSR}, and OSTrack~\cite{ostrack} (the most recent state-of-the-art from ECCV 2022).
When the tracker output is a bounding box (as most of the trackers,~\eg CSR-DCF, OSTrack), we apply AlphaRefine~\cite{alpharefine} to extract foreground segmentations (as suggested by OSTrack~\cite{ostrack}).
As in standard tracking tasks, we are given the ground truth segmentation (or bounding box) in the first frame of the video.
Then, the 2D tracker runs for the whole sequence without any re-initialization, producing a segmentation in each frame, which we feed as initialization to our method.

The proposed method processes every sequence in an online manner.
For evaluation, we compute the Intersection-over-Union (IoU) between the ground truth segmentation and our output (Sec.~\ref{sec:optim}).

As Table~\ref{tab:cdtb} shows, our approach improves segmentation performance when built on top of any of the evaluated trackers.
Average improvement is above $4 \%$ for recent methods~\cite{D3S,ostrack}.
For an older discriminative correlation filter based tracker with bounding box output, the improvement is only $2.8 \%$, probably because the initial segmentations are poor.
Many objects in this dataset are close to rigid but not perfectly rigid. 
Nevertheless, our method still successfully reconstructs such objects in motion,~\eg a human body in Fig.~\ref{fig:cdtb} and a backpack in Fig.~\ref{fig:cdtb2} (top).

\setlength{\fboxsep}{0pt}
\renewcommand{\addimg}[1]{\includegraphics[width=0.135\linewidth]{#1}}
\newcommand{\addimgframe}[1]{\framebox{\includegraphics[width=0.135\linewidth]{#1}}}

\newcommand{\makeOneRowFrame}[1]{\addimgframe{#10} & \addimgframe{#11} & \addimgframe{#12}  & \addimgframe{#13}  & \addimgframe{#14}  & \addimgframe{#15} }
\newcommand{\makeOneRow}[1]{\addimg{#10} & \addimg{#11} & \addimg{#12}  & \addimg{#13}  & \addimg{#14}  & \addimg{#15} }

\newcommand{\makeRow}[1]{%
\AlgName{\phantom{Inputt} Input} & \makeOneRow{#1s} \\
\AlgName{\phantom{Inputt} Ours} & \makeOneRowFrame{#1f} & \addimg{#1tex_deep} \\
\AlgName{\phantom{Inpu} RGB proj.} & \makeOneRowFrame{#1r} & \addimg{#1tex}\\
}

\begin{figure*}
\centering
\small
\setlength{\tabcolsep}{0.0em} 
\renewcommand{\arraystretch}{0.5} 
\begin{tabular}{@{}c@{\hskip 0.1em}ccccc@{\hskip 0.1em}c@{\hskip 1em}c@{}}

\makeRow{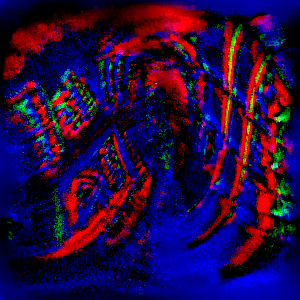}
 & \multicolumn{6}{c}{Keyframes} & Texture \\
 
\end{tabular}
\caption{\textbf{Coin tracking dataset.} The proposed method successfully tracks with 6DoF and estimates 3D shape of flat objects. Since the object in motion has different lighting effects,~\eg first and second columns, the RGB back-projection contains noisy estimates.}
\label{fig:flat}
\end{figure*}

\begin{table}
\centering
\small
\setlength{\tabcolsep}{10pt} 
\newcommand{\MySkip}{\hskip 1.3em}  
\begin{tabular}{lcc}
\toprule
 Method & IoU  & $\Delta$ \\
\midrule
Coin tracking~\cite{coin} & 0.704 & - \\
Ours (RGB feat.) & 0.718 &  2.0 \% \\
Ours (RGB feat., flat prior) &  0.731  & 3.8 \% \\
Ours (S2DNet~\cite{S2DNet} feat.) &  0.740  & 5.1 \% \\
Ours (S2DNet~\cite{S2DNet} feat., flat prior) &  \textbf{0.764}  &  \textbf{8.5 \%} \\
\bottomrule
\end{tabular}
\vspace{0.3em}
\caption{\textbf{Coin tracking dataset~\cite{coin}.} We achieve the highest segmentation accuracy with S2DNet~\cite{S2DNet} features and flat-object prior. }
\label{tab:coin}
\end{table}

\subsection{Coin Tracking dataset}
%
The Coin Tracking dataset~\cite{coin} was proposed to address a new problem, namely tracking of both sides of flat, potentially fast-rotating rigid objects  such as cards, books, and coins. 
The dataset creators also introduced a specific method to tackle the flat-object tracking problem.
Therefore, we also integrate a flat-object prior into our method by projecting all vertices to a single plane after each iteration of optimization.

When using simple RGB values as appearance features, the segmentation accuracy improves by $2\%$, and by $3.8\%$ with the flat-object prior (Table~\ref{tab:coin}).
In contrast, when using S2DNet~\cite{S2DNet} features, the improvement is $5.1\%$, and $8.5 \%$ with the flat-object prior.
The final reconstruction with the flat-object prior is shown in Fig.~\ref{fig:flat} and Fig.~\ref{fig:cdtb2} (middle row).

\begin{table}
\centering
\small
\setlength{\tabcolsep}{15pt} 
\begin{tabular}{lcc}
\toprule
 Ablated version & IoU  & $\Delta$ \\
\midrule
OSTrack~\cite{ostrack} & 0.703 & - \\
Ours (RGB feat.) & 0.717 & +2.0 \% \\
\midrule
Ours (S2DNet~\cite{S2DNet} feat.) & \textbf{0.740} & \textbf{+5.3 \%} \\
No $\Loss_{\Feat}$~\eqref{eq:feat} & 0.727 & +3.4 \%  \\
No $\Loss_M$~\eqref{eq:motion} & 0.689 & -2.0 \%  \\
No $\Loss_L$ (Sec.~\ref{sec:loss}) & 0.692 & -1.6 \%  \\
No $\Loss_T$ (Sec.~\ref{sec:loss}) & 0.734 & +4.4 \%  \\
\bottomrule
\end{tabular}
\vspace{0.3em}
\caption{\textbf{Ablation study.} Each row in the bottom part represents segmentation accuracy change w.r.t.~OSTrack~\cite{ostrack} when each of the introduced loss terms (in Sec.~\ref{sec:loss}) is disabled.}
\label{tab:ablation}
\end{table}

\subsection{In-hand object manipulation dataset}
This dataset proposed by~\cite{Wang2019} contains 13 sequences of in-hand manipulation of objects from the YCB dataset~\cite{ycb}. 
Each sequence ranges from 300 to 700 frames in length.
The ground truth is provided as manually labeled segmentation masks for every tenth frame.
We again evaluate accuracy of our reprojected silhouettes (with OSTrack initial segmentations and S2DNet features).
As shown in Table~\ref{tab:inhand}, our method brings a small improvement from $93.26 \%$ to $94.05 \%$.
We believe that the performance on this dataset is already saturated and larger improvements are hardly achievable due to inaccurate ground truth annotations.
Final reconstructions are shown in Fig.~\ref{fig:cdtb2} (two bottom rows).

\begin{figure}
\centering
\resizebox {0.4\textwidth}{!} {\begin{tikzpicture}
\begin{axis}[legend pos=north west,
    ymin=0.7, ymax=0.75,
    minor y tick num = 3,
    area style,
    xlabel={\# of keyframes},
    ylabel={IoU},
    ]

\addplot+[ybar,mark=no,cyan!40,fill opacity=0.6,] plot coordinates { 
(1, 0.7034)
(2, 0.7303)
(3, 0.7313)
(4, 0.7338)
(5, 0.7344)
(6, 0.7404)
(7, 0.7377)
(8, 0.7429)
(9, 0.7362)
(10, 0.7423)
};
\end{axis}
\end{tikzpicture}}
\vspace{-6pt}
\caption{\textbf{Influence of the number of keyframes.} When the number of keyframes is increased, the average IoU increases as well thanks to more constraints from different views.}
\label{fig:key}
\end{figure}
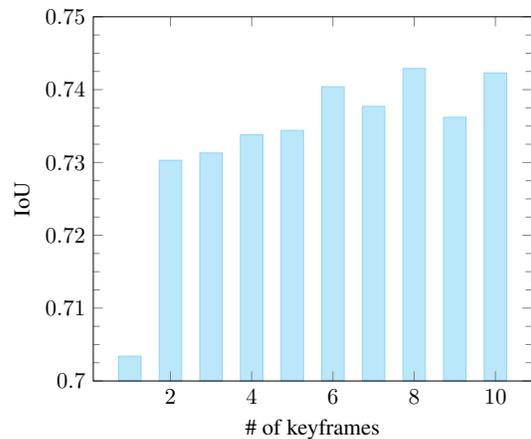

\newcommand{\addimgno}[1]{{\includegraphics[width=0.12\linewidth]{#1}}}
\renewcommand{\addimg}[1]{\framebox{\addimgno{#1}}}

\newcommand{\makeOneRoww}[1]{\addimg{#10} &\addimg{#11} & \addimg{#13}  & \addimg{#15}  & \addimg{#16}  & \addimg{#18} }
\newcommand{\makeRoww}[1]{%
\addimgno{#1i0} & \addimgno{#1i8} & \makeOneRoww{#1r} \\\addlinespace[0.1em]
}

\begin{figure*}
\centering
\small
\setlength{\tabcolsep}{0.1em} 
\renewcommand{\arraystretch}{0} 
\begin{tabular}{@{}cc@{\hskip 0.2em}cccccc@{}}

\makeRoww{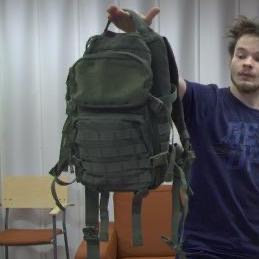}
\makeRoww{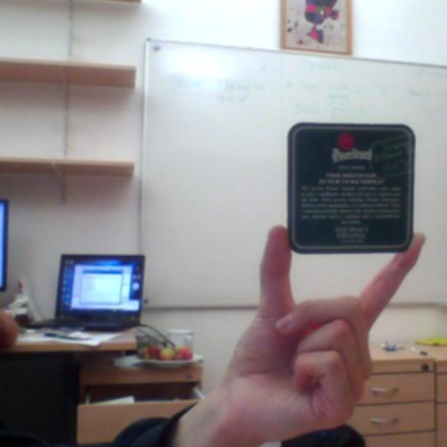}
\makeRoww{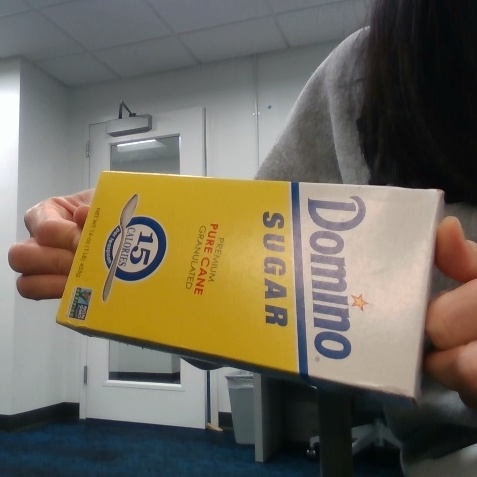}
\makeRoww{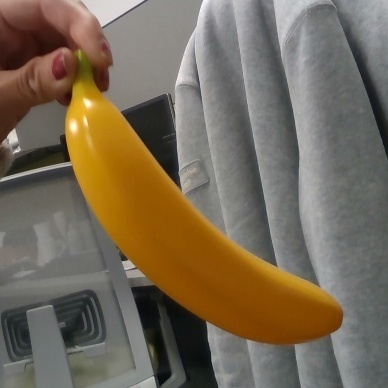}
 \multicolumn{2}{c}{Input example} & \multicolumn{6}{c}{Rendered appearance at keyframes} \\ \addlinespace[0.3em]
 
\end{tabular}
\caption{\textbf{Final 3D textured shape reconstructions.} We visualize RGB back-projections (after optmizing with S2DNet~\cite{S2DNet} features) on a backpack and container sequences from the CDTB dataset~\cite{cdtb} (top row), a beermat sequence from the coin tracking dataset~\cite{coin} (second row), and sugar and banana sequences from~\cite{Wang2019} (bottom two rows).  }
\label{fig:cdtb2}
\end{figure*}

\subsection{Ablation study}\label{sec:ablation}
%
We ablate the main components of the proposed method on 16 random sequences from the CDTB dataset~\cite{cdtb}.
As shown in Table~\ref{tab:ablation}, disabling any part of the total loss function $\Loss$ (Sec.~\ref{sec:loss}) decreases performance, indicating that all components of our method contribute to the overall accuracy.
The silhouette rendering loss~\eqref{eq:sil} cannot be disabled since in that case, the method does not know where the object of interest is and never converges.
Another interesting observation is that using RGB features ('RGB feat.' row) is worse than completely discarding the feature rendering loss ('No $\Loss_{\Feat}$ row'; $+2 \%$ vs $+3.4 \%$ improvement) since the lighting and illumination effects make the RGB features rather unreliable.
For instance, this can be observed in a vacuum cleaner example where the direct light does not move with the object (Fig.~\ref{fig:vacuum}, top row), and in a card tracking example where shadows change colors drastically in the first and second columns (Fig.~\ref{fig:flat}).
The influence of the number of keyframes $\param_K$ is shown in Fig.~\ref{fig:key}.
As expected, the more keyframes are used, the better are results, but it comes at the cost of the increased runtime.
In all experiments, we used $\param_K = 6$, which is a reasonable trade-off value.

\subsection{Benchmark for 6DoF Object Pose}\label{sec:bop}
As a by-product of the proposed tracker, we additionally produce an approximate 3D shape and 6DoF pose of the object.
To evaluate these estimates, we use the TUD-L dataset from the Benchmark for 6D Object Pose (BOP)~\cite{hodan2020bop}.
We chose this dataset because it contains videos of one moving object with limited occlusions, matching the common setting in standard 2D tracking datasets, which we target. 
As shown in Table~\ref{tab:bop}, our method produces results competitive with several recent methods,
{\em while being the only one that does not need the 3D object model as input}, neither at the training nor the test time. 
Instead, we {\em estimate it} along with the 6DoF pose.
Hence, this is a strong result.
Moreover, we do not train on the TUD-L dataset at all: our optimization-based method works out of the box, showing strong generalization across datasets.
In fact, all other compared methods are trained on large synthetic and real datasets and use various refinement steps (ICP).

\begin{table}
\centering
\footnotesize
\setlength{\tabcolsep}{6pt} 
\newcommand{\MySkip}{\hskip 1.3em}  
\begin{tabular}{lccc}
  \toprule
         &               & given    & requires\\
  Method & AR$\uparrow$  & 3D model & training\\
  \midrule
  CosyPose-ECCV20-Synt+Real       & 82.3 & yes & yes \\
  CDPNv2 BOP20-RGB                & 77.2 & yes & yes \\
  CDPN BOP19-RGB                  & 76.9 & yes & yes \\
  Zhigang-CDPN-ICCV19             & 75.7 & yes & yes \\
  Leaping from 2D to 6D-ECCVW20   & 75.1 & yes & yes \\
  CosyPose-ECCV20-PBR             & 68.5 & yes & yes \\
  CDPNv2 BOP20-PBR                & 58.8 & yes & yes \\
  EPOS-BOP20-PBR                  & 55.8 & yes & yes \\
  \textbf{Ours}                   & 54.5 & \textbf{no} & \textbf{no} \\
  Pix2Pose-BOP20-ICCV19           & 42.0 & yes & yes \\
  Sundermeyer-IJCV19              & 40.1 & yes & yes \\
  Pix2Pose-BOP19-ICCV19           & 34.9 & yes & yes \\
  SingleMultiPathEncoder-CVPR20   & 33.4 & yes & yes \\ 
  DPOD (synthetic)                & 24.2 & yes & yes\\
  \bottomrule
\end{tabular}
\vspace{0em}
\caption{\textbf{Evaluation on TUD-L dataset from the BOP benchmark~\cite{hodan2020bop}.} We compare to methods that operate on RGB images at test time. Only our method does not require a given 3D model and requires no training. AR: average recall as in~\cite{hodan2020bop}. }
\label{tab:bop}
\end{table}

\section{Conclusion}
We proposed a novel model-free tracking method that jointly estimates 3D shape, texture, and 6DoF pose of unknown rigid objects in videos, as opposed to just a 2D segmentation, which is commonly used in tracking.
As a result, the proposed tracker provides dense long-term correspondences on the object surface.
The core of our method uses differentiable rendering, which is used in visual object tracking for the first time.
The experiments demonstrated improved tracking accuracy on several tracking benchmarks.
The proposed method produces an object representation that supports diverse applications, such as augmented reality, object manipulation, and grasping. 
\blfootnote{\textbf{Acknowledgements.} This research was supported by a Google Focused Research Award and a research grant by FIFA.}

{\small
\bibliographystyle{ieee_fullname}
\bibliography{egbib}

\begin{thebibliography}{10}\itemsep=-1pt

\bibitem{osvos}
Sergi Caelles, Kevis-Kokitsi Maninis, Jordi Pont-Tuset, Laura Leal-Taix\'e,
  Daniel Cremers, and Luc {Van Gool}.
\newblock One-shot video object segmentation.
\newblock In {\em CVPR}, 2017.

\bibitem{ycb}
Berk Calli, Arjun Singh, James Bruce, Aaron Walsman, Kurt Konolige, Siddhartha
  Srinivasa, Pieter Abbeel, and Aaron~M Dollar.
\newblock Yale-cmu-berkeley dataset for robotic manipulation research.
\newblock {\em The International Journal of Robotics Research}, 36(3):261--268,
  2017.

\bibitem{dibr}
Wenzheng Chen, Jun Gao, Huan Ling, Edward Smith, Jaakko Lehtinen, Alec
  Jacobson, and Sanja Fidler.
\newblock Learning to predict 3d objects with an interpolation-based
  differentiable renderer.
\newblock In {\em NeurIPS}, 2019.

\bibitem{dibrpp}
Wenzheng Chen, Joey Litalien, Jun Gao, Zian Wang, Clement~Fuji Tsang, Sameh
  Khalis, Or Litany, and Sanja Fidler.
\newblock {DIB-R++}: Learning to predict lighting and material with a hybrid
  differentiable renderer.
\newblock In {\em NeurIPS}, 2021.

\bibitem{3dr2n2}
Christopher~B Choy, Danfei Xu, JunYoung Gwak, Kevin Chen, and Silvio Savarese.
\newblock 3d-r2n2: A unified approach for single and multi-view 3d object
  reconstruction.
\newblock In {\em Proceedings of the European Conference on Computer Vision
  ({ECCV})}, 2016.

\bibitem{blender}
Blender~Online Community.
\newblock {\em Blender - a 3D modelling and rendering package}.
\newblock Blender Foundation, Stichting Blender Foundation, Amsterdam, 2018.

\bibitem{atom}
Martin Danelljan, Goutam Bhat, Fahad~Shahbaz Khan, and Michael Felsberg.
\newblock {ATOM:} accurate tracking by overlap maximization.
\newblock In {\em {IEEE} Conference on Computer Vision and Pattern Recognition,
  {CVPR} 2019, Long Beach, CA, USA, June 16-20, 2019}, pages 4660--4669.
  Computer Vision Foundation / {IEEE}, 2019.

\bibitem{S2DNet}
Hugo Germain, Guillaume Bourmaud, and Vincent Lepetit.
\newblock S2dnet: Learning image features for accurate sparse-to-dense
  matching.
\newblock In {\em ECCV}, 2020.

\bibitem{gladkova2022directtracker}
Mariia Gladkova, Nikita Korobov, Nikolaus Demmel, Aljo{\v{s}}a O{\v{s}}ep,
  Laura Leal-Taix{\'e}, and Daniel Cremers.
\newblock Directtracker: 3d multi-object tracking using direct image alignment
  and photometric bundle adjustment.
\newblock {\em IROS}, 2022.

\bibitem{kcf}
João~F. Henriques, Rui Caseiro, Pedro Martins, and Jorge Batista.
\newblock High-speed tracking with kernelized correlation filters.
\newblock {\em IEEE Transactions on Pattern Analysis and Machine Intelligence},
  37(3):583--596, 2015.

\bibitem{epos}
Tom{\'a}{\v{s}} Hoda{\v{n}}, D{\'a}niel Bar{\'a}th, and Ji{\v{r}}{\'i} Matas.
\newblock {EPOS}: Estimating {6D} pose of objects with symmetries.
\newblock {\em CVPR}, 2020.

\bibitem{hodan2018bop}
Tom{\'a}{\v{s}} Hoda{\v{n}}, Frank Michel, Eric Brachmann, Wadim Kehl, Anders
  Glent~Buch, Dirk Kraft, Bertram Drost, Joel Vidal, Stephan Ihrke, Xenophon
  Zabulis, Caner Sahin, Fabian Manhardt, Federico Tombari, Tae-Kyun Kim,
  Ji{\v{r}}{\'i} Matas, and Carsten Rother.
\newblock {BOP}: Benchmark for {6D} object pose estimation.
\newblock {\em ECCV}, 2018.

\bibitem{hodan2020bop}
Tom{\'a}{\v{s}} Hoda{\v{n}}, Martin Sundermeyer, Bertram Drost, Yann Labb{\'e},
  Eric Brachmann, Frank Michel, Carsten Rother, and Ji{\v{r}}{\'i} Matas.
\newblock {BOP} challenge 2020 on {6D} object localization.
\newblock {\em ECCVW}, 2020.

\bibitem{got10k}
Lianghua Huang, Xin Zhao, and Kaiqi Huang.
\newblock Got-10k: A large high-diversity benchmark for generic object tracking
  in the wild.
\newblock {\em IEEE TPAMI}, 43(5):1562--1577, 2021.

\bibitem{Huang_2020_CVPR}
Zeng Huang, Yuanlu Xu, Christoph Lassner, Hao Li, and Tony Tung.
\newblock Arch: Animatable reconstruction of clothed humans.
\newblock In {\em Proceedings of the IEEE/CVF Conference on Computer Vision and
  Pattern Recognition (CVPR)}, June 2020.

\bibitem{Kart_2019_CVPR}
Ugur Kart, Alan Lukezic, Matej Kristan, Joni-Kristian Kamarainen, and Jiri
  Matas.
\newblock Object tracking by reconstruction with view-specific discriminative
  correlation filters.
\newblock In {\em CVPR}, June 2019.

\bibitem{adam}
Diederik~P. Kingma and Jimmy Ba.
\newblock Adam: {A} method for stochastic optimization.
\newblock In Yoshua Bengio and Yann LeCun, editors, {\em ICLR}, 2015.

\bibitem{Kristan2022a}
Matej Kristan, Ales Leonardis, Jiri Matas, Michael Felsberg, Roman Pflugfelder,
  Joni-Kristian Kamarainen, Hyung~Jin Chang, Martin Danelljan, Luka \v{C}ehovin
  Zajc, Alan Luke\v{z}i\v{c}, Ondrej Drbohlav, Johanna Bjorklund, Yushan Zhang,
  Zhongqun Zhang, Song Yan, Wenyan Yang, Dingding Cai, Christoph Mayer, and
  Gustavo Fernandez.
\newblock The tenth visual object tracking vot2022 challenge results, 2022.

\bibitem{vot2020}
Matej Kristan, Ales Leonardis, Ji\v{r}\'{i} Matas, Michael Felsberg, Roman
  Pflugfelder, Joni-Kristian Kamarainen, Luka \v{C}ehovin Zajc, Martin
  Danelljan, Alan Lukezic, Ondrej Drbohlav, Linbo He, Yushan Zhang, Song Yan,
  Jinyu Yang, Gustavo Fernandez, and et al.
\newblock The eighth visual object tracking vot2020 challenge results, 2020.

\bibitem{vot2021}
Matej Kristan, Ji\v{r}\'{i} Matas, Ale\v{s} Leonardis, Michael Felsberg, Roman
  Pflugfelder, Joni-Kristian Kamarainen, Hyung~Jin Chang, Martin Danelljan,
  Luka \v{C}ehovin Zajc, Alan Luke\v{z}i\v{c}, Ondrej Drbohlav, Jani Kapyla,
  Gustav Hager, Song Yan, Jinyu Yang, Zhongqun Zhang, Gustavo Fernandez, and
  et. al.
\newblock The ninth visual object tracking vot2021 challenge results, 2021.

\bibitem{VOT_TPAMI}
Matej Kristan, Ji\v{r}\'{i} Matas, Ale\v{s} Leonardis, Tomas VoJi\v{r}, Roman
  Pflugfelder, Gustavo Fernandez, Georg Nebehay, Fatih Porikli, and Luka
  \v{C}ehovin.
\newblock A novel performance evaluation methodology for single-target
  trackers.
\newblock {\em IEEE TPAMI}, 38(11):2137--2155, Nov 2016.

\bibitem{labbe2020}
Y. {Labbe}, J. {Carpentier}, M. {Aubry}, and J. {Sivic}.
\newblock Cosypose: Consistent multi-view multi-object 6d pose estimation.
\newblock In {\em ECCV}, 2020.

\bibitem{Leeb2019MotionNets6T}
Felix Leeb, Arunkumar Byravan, and Dieter Fox.
\newblock Motion-nets: 6d tracking of unknown objects in unseen environments
  using rgb.
\newblock {\em IROS Workshop on The Importance of Uncertainty in Deep Learning
  for Robotics}, 2019.

\bibitem{barf}
Chen-Hsuan Lin, Wei-Chiu Ma, Antonio Torralba, and Simon Lucey.
\newblock Barf: Bundle-adjusting neural radiance fields.
\newblock In {\em ICCV}, 2021.

\bibitem{pixsfm}
Philipp Lindenberger, Paul-Edouard Sarlin, Viktor Larsson, and Marc Pollefeys.
\newblock {Pixel-Perfect Structure-from-Motion with Featuremetric Refinement}.
\newblock In {\em ICCV}, 2021.

\bibitem{lipson2022coupled}
Lahav Lipson, Zachary Teed, Ankit Goyal, and Jia Deng.
\newblock Coupled iterative refinement for 6d multi-object pose estimation.
\newblock In {\em CVPR}, 2022.

\bibitem{softras}
Shichen Liu, Tianye Li, Weikai Chen, and Hao Li.
\newblock Soft rasterizer: A differentiable renderer for image-based 3d
  reasoning.
\newblock {\em ICCV}, Oct 2019.

\bibitem{liu2020dist}
Shaohui Liu, Yinda Zhang, Songyou Peng, Boxin Shi, Marc Pollefeys, and Zhaopeng
  Cui.
\newblock Dist: Rendering deep implicit signed distance function with
  differentiable sphere tracing.
\newblock In {\em CVPR}, 2020.

\bibitem{smpl}
Matthew Loper, Naureen Mahmood, Javier Romero, Gerard Pons-Moll, and Michael~J.
  Black.
\newblock {SMPL}: A skinned multi-person linear model.
\newblock {\em ACM Trans. Graphics (Proc. SIGGRAPH Asia)}, 34(6):248:1--248:16,
  Oct. 2015.

\bibitem{klt}
Bruce~D. Lucas and Takeo Kanade.
\newblock An iterative image registration technique with an application to
  stereo vision.
\newblock In {\em Proceedings of the 7th International Joint Conference on
  Artificial Intelligence - Volume 2}, IJCAI'81, page 674–679, San Francisco,
  CA, USA, 1981. Morgan Kaufmann Publishers Inc.

\bibitem{cdtb}
Alan Lukezic, Ugur Kart, Jani Käpylä, Ahmed Durmush, Joni-Kristian
  Kamarainen, Ji\v{r}\'{i} Matas, and Matej Kristan.
\newblock Cdtb: A color and depth visual object tracking dataset and benchmark.
\newblock In {\em ICCV}, pages 10012--10021, 2019.

\bibitem{D3S}
Alan Lukezic, Ji\v{r}\'{i} Matas, and Matej Kristan.
\newblock D3s - a discriminative single shot segmentation tracker.
\newblock In {\em CVPR}, 2020.

\bibitem{CSR}
Alan Lukezic, Tomas Voji\v{r}, Luka Cehovin~Zajc, Ji\v{r}\'{i} Matas, and Matej
  Kristan.
\newblock Discriminative correlation filter with channel and spatial
  reliability.
\newblock In {\em CVPR}, July 2017.

\bibitem{nerf}
Ben Mildenhall, Pratul~P. Srinivasan, Matthew Tancik, Jonathan~T. Barron, Ravi
  Ramamoorthi, and Ren Ng.
\newblock Nerf: Representing scenes as neural radiance fields for view
  synthesis.
\newblock In {\em ECCV}, 2020.

\bibitem{7139520}
Alessandro Pieropan, Niklas Bergström, Masatoshi Ishikawa, and Hedvig
  Kjellström.
\newblock Robust 3d tracking of unknown objects.
\newblock In {\em ICRA}, pages 2410--2417, 2015.

\bibitem{sharf}
Konstantinos Rematas, Ricardo Martin-Brualla, and Vittorio Ferrari.
\newblock Sharf: Shape-conditioned radiance fields from a single view.
\newblock In {\em ICML}, 2021.

\bibitem{sfb}
Denys Rozumnyi, Martin~R. Oswald, Vittorio Ferrari, and Marc Pollefeys.
\newblock Shape from blur: Recovering textured 3d shape and motion of fast
  moving objects.
\newblock In {\em NeurIPS}, 2021.

\bibitem{mfb}
Denys Rozumnyi, Martin~R. Oswald, Vittorio Ferrari, and Marc Pollefeys.
\newblock Motion-from-blur: 3d shape and motion estimation of motion-blurred
  objects in videos.
\newblock In {\em CVPR}, Jun 2022.

\bibitem{schoenberger2016sfm}
Johannes~Lutz Sch\"{o}nberger and Jan-Michael Frahm.
\newblock Structure-from-motion revisited.
\newblock In {\em Conference on Computer Vision and Pattern Recognition
  (CVPR)}, 2016.

\bibitem{klt2}
Jianbo Shi and Tomasi.
\newblock Good features to track.
\newblock In {\em 1994 Proceedings of IEEE Conference on Computer Vision and
  Pattern Recognition}, pages 593--600, 1994.

\bibitem{coin}
Jon\'{a}\v{s} \v{S}er\'{y}ch and Ji\v{r}\'{\i} Matas.
\newblock Visual coin-tracking: Tracking of planar double-sided objects.
\newblock In {\em GCPR}, page 317–330, Berlin, Heidelberg, 2019.
  Springer-Verlag.

\bibitem{Wang_2021_ICCV}
Dan Wang, Xinrui Cui, Xun Chen, Zhengxia Zou, Tianyang Shi, Septimiu Salcudean,
  Z.~Jane Wang, and Rabab Ward.
\newblock Multi-view 3d reconstruction with transformers.
\newblock In {\em ICCV}, pages 5722--5731, October 2021.

\bibitem{Wang2019}
Fan Wang and Kris Hauser.
\newblock In-hand object scanning via rgb-d video segmentation.
\newblock {\em ICRA}, 2019.

\bibitem{wang2020self6d}
Gu Wang, Fabian Manhardt, Jianzhun Shao, Xiangyang Ji, Nassir Navab, and
  Federico Tombari.
\newblock Self6d: Self-supervised monocular 6d object pose estimation.
\newblock In {\em The European Conference on Computer Vision (ECCV)}, August
  2020.

\bibitem{alpharefine}
Bin Yan, Xinyu Zhang, Dong Wang, Huchuan Lu, and Xiaoyun Yang.
\newblock Alpha-refine: Boosting tracking performance by precise bounding box
  estimation.
\newblock In {\em CVPR}, pages 5285--5294, 2021.

\bibitem{ostrack}
Botao Ye, Hong Chang, Bingpeng Ma, Shiguang Shan, and Xilin Chen.
\newblock Joint feature learning and relation modeling for tracking: A
  one-stream framework.
\newblock In {\em ECCV}, 2022.

\bibitem{inerf}
Lin Yen-Chen, Pete Florence, Jonathan~T. Barron, Alberto Rodriguez, Phillip
  Isola, and Tsung-Yi Lin.
\newblock {iNeRF}: Inverting neural radiance fields for pose estimation.
\newblock {\em IROS}, 2020.

\bibitem{you2022cppf}
Yang You, Ruoxi Shi, Weiming Wang, and Cewu Lu.
\newblock Cppf: Towards robust category-level 9d pose estimation in the wild.
\newblock In {\em CVPR}, 2022.

\bibitem{pixelnerf}
Alex Yu, Vickie Ye, Matthew Tancik, and Angjoo Kanazawa.
\newblock {pixelNeRF}: Neural radiance fields from one or few images.
\newblock In {\em CVPR}, 2021.

\end{thebibliography}
}

\end{document}